\documentclass[lettersize,journal]{IEEEtran}
\usepackage{graphicx}
\usepackage{amsmath}
\usepackage{amssymb}
\usepackage{times}
\usepackage{soul}
\usepackage{epsfig}
\usepackage{amsthm}
\usepackage{amsfonts}
\usepackage{bbm}
\usepackage{booktabs}
\usepackage{color}
\usepackage{multirow}
\usepackage{array}
\usepackage{enumitem}
\usepackage{amsmath,amsfonts}
\usepackage{algorithmic}
\usepackage{algorithm}
\usepackage{array}
\usepackage{subfig}
\usepackage{textcomp}
\usepackage{stfloats}
\usepackage{url}
\usepackage{verbatim}
\usepackage{graphicx}
\usepackage{cite}

\begin{document}

\title{Learning Unsupervised Gaze Representation via Eye Mask Driven Information Bottleneck}

\author{Yangzhou Jiang$^*$, Yinxin Lin$^*$, Yaoming Wang, Teng Li, Bilian Ke$^\dagger$, Bingbing Ni$^\dagger$
\thanks{Authors are with Shanghai Jiao Tong University.}
\thanks{$*$ indicates equal contribution.}
\thanks{$\dagger$ indicates joint corresponding authors.}
}

\markboth{IEEE Transactions on Image Processing,~Vol.~XX, No.~XX, August~2024}%
{Shell \MakeLowercase{\textit{et al.}}: A Sample Article Using IEEEtran.cls for IEEE Journals}


\maketitle

\begin{abstract}
Appearance-based supervised methods with full-face image input have made tremendous advances in recent gaze estimation tasks. However, intensive human annotation requirement inhibits current methods from achieving industrial level accuracy and robustness.
Although current unsupervised pre-training frameworks have achieved success in many image recognition tasks, due to the deep coupling between facial and eye features, such frameworks are still deficient in extracting useful gaze features from full-face.
To alleviate above limitations, this work proposes a novel unsupervised/self-supervised gaze pre-training framework, which forces the full-face branch to learn a low dimensional gaze embedding without gaze annotations, through collaborative feature contrast and squeeze modules. In the heart of this framework is an alternating eye-attended/unattended masking training scheme, which squeezes gaze-related information from full-face branch into an eye-masked auto-encoder through an injection bottleneck design that successfully encourages the model to pays more attention to gaze direction rather than facial textures only, while still adopting the eye self-reconstruction objective. In the same time, a novel eye/gaze-related information contrastive loss has been designed to further boost the learned representation by forcing the model to focus on eye-centered regions.
Extensive experimental results on several gaze benchmarks demonstrate that the proposed scheme achieves superior performances over unsupervised state-of-the-art.

\end{abstract}


\section{Introduction}
\IEEEPARstart{G}{aze} estimation lays a foundation for many applications including immersive human-computer interaction and behavior analysis, thus attracting increasing research attention, especially in the coming era of meta-verse~\cite{li2022appearance, yucel2013joint, topal2013low, zhang2019scene, sun2017camera}.
Current widely used appearance-based gaze estimation methods, usually retrieve gaze direction from monocular full-face images via neural networks exhibiting good performance with the development of advanced deep learning architectures.
However, these approaches depend heavily on intensive and precise annotation acquisition, which is cost-consuming.
Moreover, suffering from the domain gap, models learned from large-scale labeled gaze datasets degrade dramatically
when transferred to an unseen application domain (with different background and illumination conditions).
To alleviate above problems, some recent works~\cite{yu2020unsupervised, sun2021cross} attempt to learn gaze direction in an unsupervised manner, \emph{i.e.}, using eye patch reconstruction (self-reconstruction) with CNN based auto-encoders~\cite{he2022masked}.
However, the performance of these approaches still lags far behind that of supervised approaches, given that a good trade-off between the reconstruction accuracy of the eye region and the full-face region is hard to obtain using existing training schemes.
In addition, they have difficulty in dealing with the great variation of head poses.
\begin{figure}[!t]
    \centering
    \includegraphics[width=1.0\linewidth]{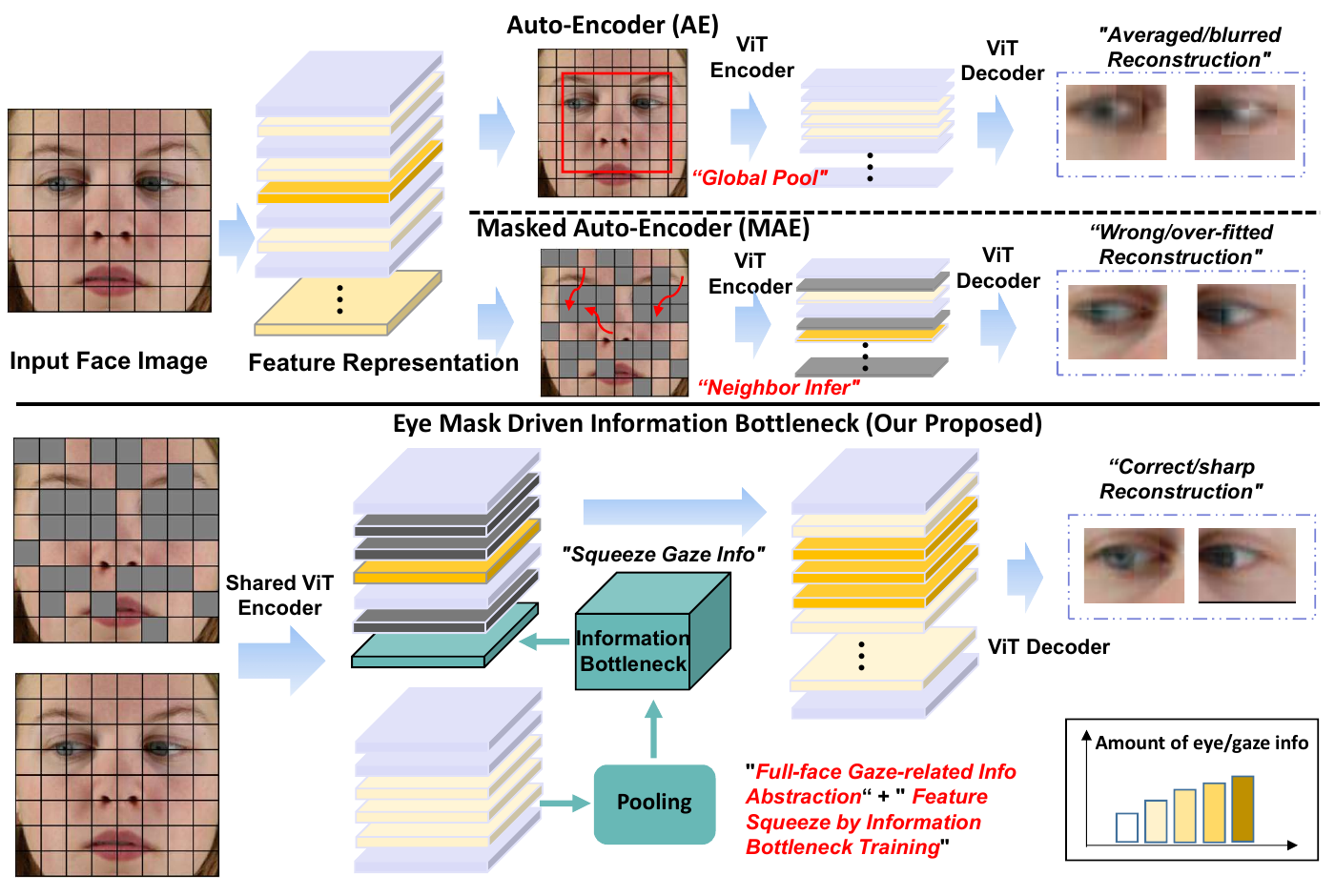}
    \caption{Motivation of the proposed work. Masked Auto-Encoder (MAE)~\cite{he2022masked} learns by completing highly masked images (and eyes are most probably masked out), which highly possibly yields over-fitting eye/gaze representations. Auto-Encoder (AE)~\cite{vincent2008extracting} learns from whole image dimensionality reduction and reconstruction, and eye/gaze information is attenuated during full-face feature learning and pooling. Thus, both AE and MAE tend to learn general face information (or even non-eye information such as facial texture) instead of focus on gaze information.
    In contrast, our proposed Eye Mask Driven Information Bottleneck (EM-IB) unsupervised learning scheme, can successfully enforce the model to concentrate on eye-gaze related information by 1) a full-face global feature injection via a novel information bottleneck structure design and 2) a newly proposed eye/gaze information contrastive training loss.}
    \label{fig:motivation}
\end{figure}

Recent fast development of novel unsupervised learning schemes have shed some light for unsupervised gaze estimation. For classification tasks, the paradigms of \textbf{contrastive}~\cite{dino,simclr,chen2020big,chen2020improved,chen2021exploring,he2020momentum} and \textbf{generative} learning~\cite{bao2021beit,chen2020generative,he2022masked,zhou2022image,xie2021simmim} have been prevailing the recent progress of self-supervised learning and exhibiting even more competitive performance compared to supervised methods.
Despite promising results in several tasks, contrastive learning methods are not suitable for unsupervised gaze estimation.
On the one hand, different views of a single face image focus on global semantic information that benefits classification tasks, ~\emph{e.g.}, information for face recognition, detection or parsing, but local regions around eyes are not well-attended, therefore insufficient gaze-related features could be extracted for gaze-estimation.
On the other hand, since the gaze direction is a two-dimensional manifold and lacks rotation equivalence, no suitable augmentations like cutmix~\cite{yun2019cutmix}, multi-crop~\cite{caron2020unsupervised} can be employed to create different views from a single image.
Compared with contrastive methods, generative methods do not need delicate design of multiple views, and they
learn the latent representation by recovering densely masked images in pixel level~\cite{he2022masked} or token level~\cite{bao2021beit} with vision transformers~\cite{dosovitskiy2020image} (ViT).
However, existing methods such as reconstructing pixels with a random mask like MAEs~\cite{he2022masked} still can not accommodate to gaze estimation task for following reasons (please refer to Fig.~\ref{fig:motivation} for an intuitive understanding).
First, gaze-direction related information is mostly centered around eye regions occupying only small parts in the face, thus if the eye region is masked the model could hardly infer the gaze direction (\emph{i.e.}, which might mislead the model to learn over-fitted feature representation). Second, methods compressing and reconstructing images like AEs~\cite{vincent2008extracting} might induce a large portion of redundant or noisy features, since reconstructing eye-related pixels from the face input most possibly focuses the model more on full-face semantic information, which in turn leads to gaze estimation ambiguity. \textbf{As a summery, there lacks a principled way to extract (push) more gaze-related feature from the full-face image which can be then injected to reconstruct the masked eye region, \emph{i.e.}, with sufficient gaze direction information}.

To address above challenges, we propose in this work a novel unsupervised gaze estimation framework, called \textbf{Eye Mask Driven Information Bottleneck (EM-IB)}, to guide the full-face feature extraction branch to learn a low dimensional gaze-related information embedding while avoiding conveying too much facial semantic information (not helpful for gaze representation).
The proposed framework features the following designs.
First, a dual-branch gaze information extraction and squeeze architecture is proposed composing of an Eye-masked Auto-Encoder (EM-AE) branch as well as a Full-face Information Injection Bottleneck (FF-IB) branch, by which, gaze related representation could be successfully captured/abstracted/distilled out of a large portion of facial patches in the FF-IB branch and injected/self-squeezed to the EM-AE branch through an information bottleneck structure, via training with the eye patch prediction and reconstruction objective. By this self-squeeze design, it is able to force the injection bottleneck to encode more head pose/eye gaze related information while ensuring the reconstruction still pays attention to the gaze direction rather than the facial textures.
In addition, we observe that the region of the face can be divided into eye regions and non-eye regions, and the visual features extracted from these different regions contribute different information to the prediction of gaze. For example, according to the orientation/position of the pupil in the eye, the eye region provides the richest information of gaze direction. On the contrary, areas outside the eyes, although they also contain head pose information, contain many facial details such as wrinkles and muscle movements which can even confuse the gaze-related feature representation.
The above fact motivates us with a new self-supervision strategy besides the conventional self-reconstruction loss. Specifically, we construct a novel triplet based auxiliary unsupervised training loss (called \textbf{Unsupervised Eye/gaze Information Contrastive Loss}), which constrains that the reconstruction MSE error computed from more gaze related information injection (full-face plus eye patches) to be smaller than that of less gaze information injection (full-face with non-eye patches). This novel training objective in conjunction with the eye reconstruction loss further guide the full-face branch to learn better feature representation for gaze direction estimation.

Our proposed unsupervised appearance-based gaze estimation framework achieves comparable performance of supervised methods with only $100$ samples of linear probing.
Experimental results demonstrate that our proposed pre-training scheme (\emph{i.e.}, EM-IB) significantly surpasses the state-of-the-arts in both unsupervised and domain adaptation settings.
Specifically, EM-IB achieves a $29.2\%$ performance improvement compared with the state-of-the-art unsupervised gaze estimation approach cross-encoder (CE)~\cite{sun2021cross} on MPIIFaceGaze.
For 100-shot cross dataset experiments, EM-IB without labels of the source data even outperforms the supervised method trained with labeled source data on Gaze360. Besides, we show our pre-trained injection encoder can be distilled into a lightweight model, e.g., ResNet18 distills from ViT without performance degradation.

\section{Related Works}
\subsection{Appearance-based Gaze Estimation}
Appearance-based approaches learn image-to-gaze mapping with images captured by remote camera directly, and this work focuses on unsupervised appearance-based gaze learning. In early works of appearance-based gaze estimation with deep models, researchers attempt to regress gaze from cropped eye patches with different convolution neural networks, \emph{e.g.}, LeNet~\cite{zhang2015appearance}, AlexNet~\cite{zhang2017s} and VGGNet~\cite{zhang2017mpiigaze}. The input of the network is the cropped eye patch, and the output of the network is the gaze vector, which includes two components: pitch and yaw angles.
It has been discovered that incorporating facial information is also beneficial for gaze estimation, leading many methods to attempt the fusion of full facial features into neural networks~\cite{gazecapture, cheng2020coarse, chen2018appearance}.

Krafka et al.~\cite{gazecapture} utilize a triple-path network to extract features from two eyes patch and face patch. Zhang et al.~\cite{zhang2017s} take a full-face image as input, and compute spatial weights on feature map to enhance information on eye region to learn the gaze direction. Work~\cite{cheng2020coarse} uses a coarse-to-fine strategy to predict gaze from full-face input and gaze residual from eye patches input respectively.
Besides, regressing gaze from full-face input directly could lead to comparable performance.
Zhang et al.~\cite{zhang2020eth} achieve comparable performance by training a ResNet-50~\cite{he2016deep} to regress gaze from a full-face image.
We recommend the readers refer to an excellent survey~\cite{cheng2021appearance} for a comprehensive understanding of appearance-based gaze estimation methods.
\subsection{Unsupervised Gaze Estimation}
To ease the difficulty of obtaining excessive gaze annotations, some recent works resort to unsupervised learning solutions.
Previous unsupervised attempts learn the latent gaze representation mainly by reconstructing eye patch with auto-encoders.
Yu et al.~\cite{yu2020unsupervised} utilize a gaze redirection network for unsupervised gaze representation learning. They take the latent representation difference of the pair of images from the same person with the same head pose as a redirection variable, which makes two latent variables linearly related to pitch and yaw angles. Sun et al.~\cite{sun2021cross} propose using cross-encoder to disentangle the features with a latent-code-swapping mechanism on eye-consistent image pairs and gaze-similar ones. Gideon et al.~\cite{Gideon_2022_cvpr} expand their work to multiple synchronized views of a person's face, and learn gaze representation with a cross-encoder from unlabeled multi-view video dataset.
These auto-encoder methods only learn unsupervised gaze representation from cropped eye patches and ignore the rich information of the whole human face, which is beneficial for gaze estimation as suggested in~\cite{zhang2017s}.
Our unsupervised solution also admits full-face input and could benefit from larger full-face datasets.
Except for these reconstruction methods, work~\cite{kothari2021weakly} leverages weak supervision from external videos of human interactions to boost the gaze estimation performance in semi-supervised and cross-domain generalization setting on physically unconstrained dataset Gaze360~\cite{kellnhofer2019gaze360}. Besides, there are methods enhance the domain generalization and adaptation performance of gaze estimation in full-face input gaze setting~\cite{cheng2022puregaze, liu2021generalizing, Wang_2022_cvpr, Bao_2022_cvpr}. Note that we do not study the gaze domain adaptation problem in this work.

\vspace{-0.2cm}
\section{Methodology}
\begin{figure*}[htb]
    \centering
    \includegraphics[width=1.0\linewidth]{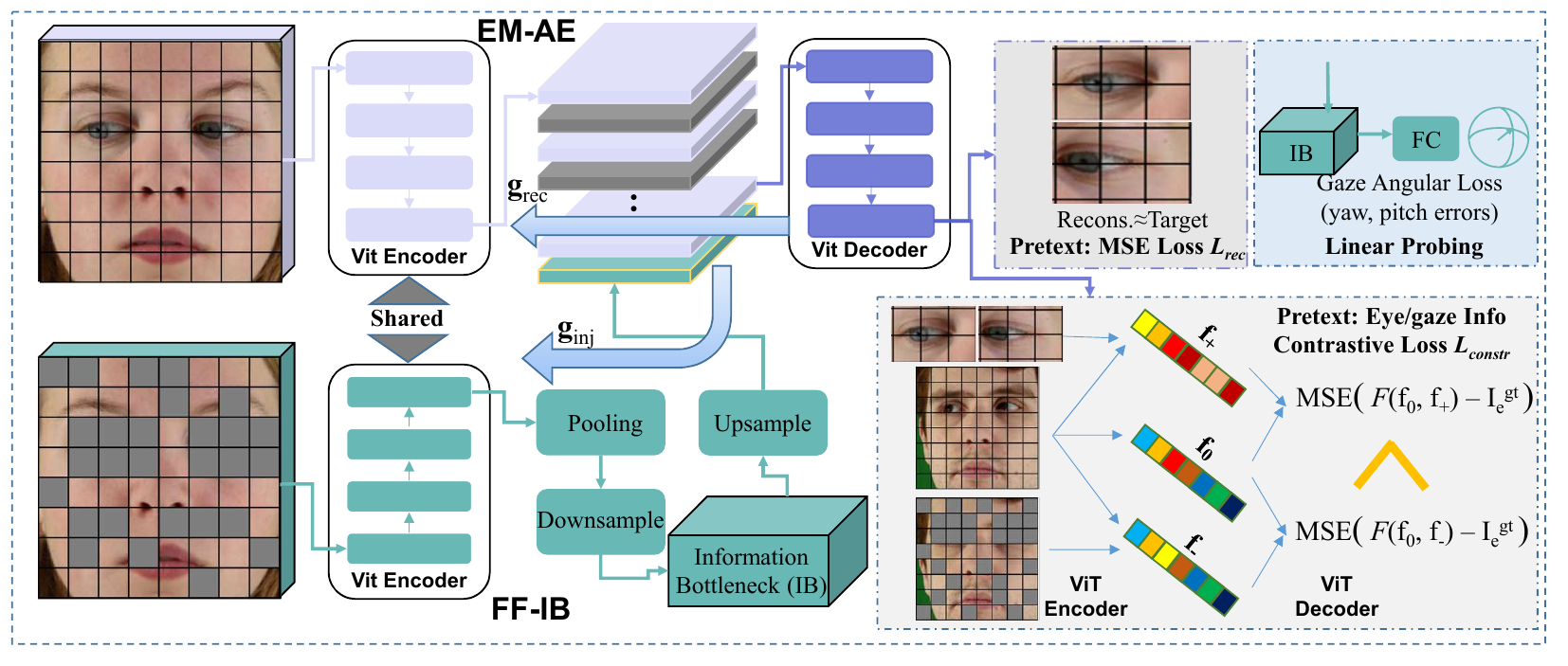}
    \caption{Illustration of our proposed Eye Mask Driven Injection Bottleneck (EM-IB) unsupervised gaze learning pipeline. The upper branch is the Eye-masked Auto-Encoders (EM-AE), which extracts eye/gaze-related information based on the unmasked patches to reconstruct the masked eye-area. The bottom branch is the Full-face Injection Bottleneck (FF-IB), which injects a compressed full-face level gaze-related vector to the EM-AE module. The eye/gaze information self-squeeze is achieved by this asymmetric encoder-decoder structure. During unsupervised pre-training phase, both MSE loss and eye/gaze information contrastive loss are utilized. For linear probing in gaze estimator training phase, conventional 2D gaze angular loss is utilized. Gradient flows (within the weight sharing ViT encoder structure) for both reconstruction and injection branches are also indicated.
    }
    \label{fig:framework}
\end{figure*}

We introduce in detail our proposed unsupervised pre-training framework for gaze estimation in this section, \emph{i.e.}, Eye Mask Driven Injection Bottleneck (EM-IB). To this end, we first revisit a popular un/self-supervised training scheme, namely, masked auto-encoder~\cite{he2022masked} and point out its inability in extracting useful gaze-related representation in our task. To address this challenge, we then introduce our proposed novel pre-training scheme which masks out eye and non-eye regions alternately forcing the full-face feature extractor to learn gaze-related representation and squeeze this information towards better eye reconstruction via an information bottleneck structure and a novel eye/gaze information contrastive training loss design. Fig.~\ref{fig:framework} illustrates the pipeline structure of the proposed framework.


\subsection{Preliminary: Masked Auto-Encoder of ViT for Unsupervised Learning}
\label{sec:preliminary}
Masked Auto-Encoder (MAE)~\cite{he2022masked} exhibits remarkable performance by reconstructing pixels directly with ViT~\cite{dosovitskiy2020image} from highly masked images. Like all auto-encoders, MAE contains an encoder to map input image to latent representation and a decoder to reconstruct pixels from the latent representation. Unlike CNN that operates on regular grids, ViT takes patches with position embedding as the input, which is more flexible. During unsupervised pre-training, MAE randomly masks input patches which situate arbitrarily on the entire image and takes the rest patches as the input for the encoder. Then the decoder takes the encoded token and masked token with position embedding to reconstruct the masked pixel.
Besides, MAE suggests masking a very high portion of random patches to force the ViT to learn semantic information instead of simply copying pixels.

Despite the success of MAE on classification, direct replication on gaze estimation causes serious problems.
Specifically, reconstructing pixels with a random mask like MAEs focuses the model more on higher level face semantic information, and thus this scheme offers no meaningful information for gaze estimation task, \emph{i.e.}, \textbf{too coarse or too global facial information with no respect to gaze direction}.
However, gaze direction is closely related to the slight rotation of human eyes and heads and this information is hidden in the entire facial image pattern, which imposes the requirement that a good feature extractor should delicately encode the head/eye pose related information from the full-face image.
In the meantime, if we completely mask out the pixels associated with the eyes and perform the eye reconstruction task, no sufficient information about gaze directions can be learned leading to totally non-sense (\emph{i.e.}, over-fitting) gaze predictions, as illustrated in Fig.~\ref{fig:motivation}.
The above dilemma (\emph{i.e.}, must extract head/eye pose related information from the full-face image while must avoid encoding too much higher level facial semantic information that might contaminate the gaze-related feature representation) makes the design of unsupervised pre-training task for gaze estimation very challenging.
\subsection{Eye Mask Driven Information Bottleneck}
\label{sec:eye-mask-AE}
\begin{figure}[!t]
\renewcommand{\baselinestretch}{1.0}
    \centering
    \includegraphics[width=1.0\linewidth]{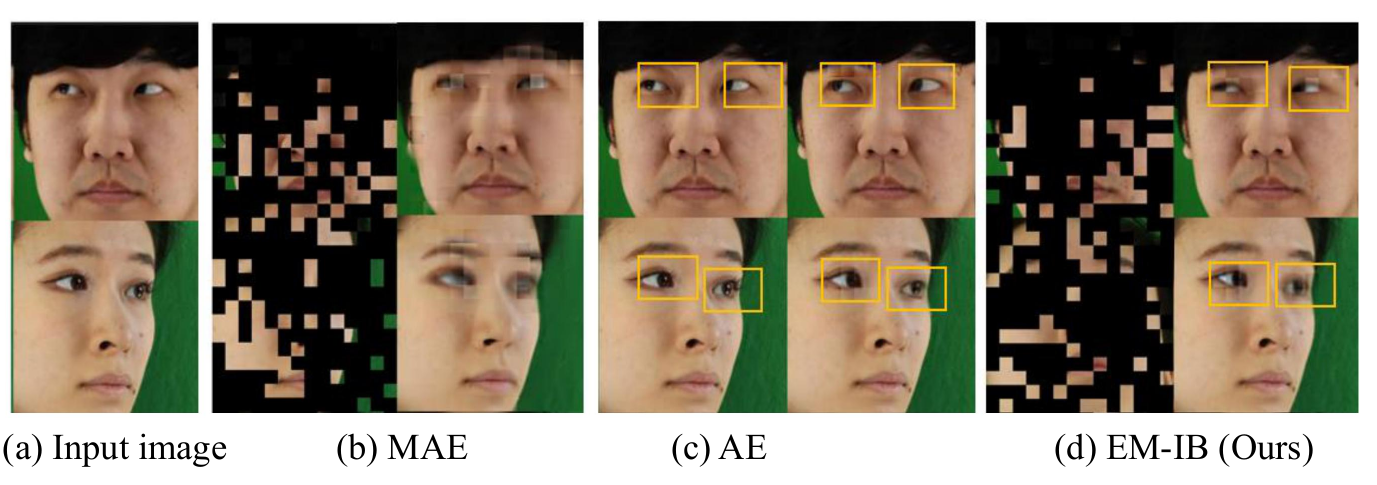}
    \caption{Visualization of the reconstructed eyes from masked image input. We show the ground-truth (a) and the masked faces and eye patches (b)-(d) reconstructed by MAE (b), AE (c) and our proposed EM-IB (d), respectively.
    }
    \label{fig:recons}
\end{figure}

To inherit the unsupervised learning power of MAE as well as to address its inability in gaze-related tasks, we propose a novel eye mask driven information bottleneck training scheme with the inspiration illustrated in Fig.~\ref{fig:motivation}.
As shown in Fig.~\ref{fig:motivation}, vanilla MAEs can hardly reconstruct the masked eye pixels, as vanilla MAEs with random mask patches over the whole image attempt to learn more global semantic information instead of gaze-relevant information.
Therefore, in order to encourage the model to learn more eye-related representation (which most probably contains gaze-related information), in stead of random mask-out, we might mask out eye related patches and train the model to reconstruct these eye regions.
Not surprisingly, the network could often \emph{guess} the right gaze direction with eyes masked full facial image, as visualized in Fig.~\ref{fig:recons}. This indicates that there is complementary gaze direction related information contained in the whole facial pattern, and it is therefore beneficial to taking the full-face image for gaze estimation, which further supports our above idea, \emph{i.e.}, encoding full-face image for assisting eye reconstruction during unsupervised pre-training.
In the following, we will introduce in detail the proposed network architecture for both full-face and eye-masked gaze-related feature extraction as well as our novel training objectives to dig out gaze-related information from full-face image out of dominating facial semantic features, in the unsupervised eye reconstruction setting.


%

\textbf{Injection Bottleneck Structure.}
To deal with above mentioned challenges, we design a novel pretext learning task which leverages an injection bottleneck structure to extract gaze-related information from unmasked full-face images (decoupled from the dominant/redundant facial semantic information) and inject it into an eye-masked auto-encoder structure as an extra token for the decoder to well predict/reconstruct the masked eye pixels.
The proposed Injection Bottleneck (IB) structure contains two major components, including 1) an eye-masked feature encoder and decoder branch (EM-AE); and 2) a full-face feature encoding and information injection bottleneck structure which squeezes full-face encoding into the eye-masked decoder for eye reconstruction (FF-IB).

For both eye-masked and full-face feature extraction branches, we employ vision transformers as the encoder and decoder structure, similar to that of MAE. Note that we employ two kinds of ViT including the base version~\cite{dosovitskiy2020image} and a tiny version~\cite{pmlr-v139-touvron21a} respectively in our work. In particular, when the base version of ViT (ViT-base) is used as encoder, it is composed by $12$ self-attention layers with the embedding dimension of $768$ and the heads number of $12$. The decoder is composed by $4$ self-attention layers with embedding dimension of $768$ and heads number of $6$. For ViT-tiny, the encoder is composed by $12$ self-attention layers with the embedding dimension as $192$ and the heads number as $3$. The decoder is composed by $4$ self-attention layers with the same embedding dimension and heads number as encoder.
As mentioned, the original sine-cosine version positional encoding scheme is used~\cite{vaswani2017attention}. Note that both branches share exactly the same structure and the weights. (The reason for weight sharing would be elaborated in the following.)
The purpose of the full-face branch is to mine and condense the high-dimensional facial feature into a low-dimensional gaze-related representation at a global/full-face level, \emph{i.e.}, and with our delicately designed training scheme, to avoid simply learning eye-pixel features such as textures (or even directly copying pixels from the eye region), for masked eye reconstruction.
Accordingly, during the pre-training phase, the eye-masked branch masks out eye-surrounded patches (\emph{i.e.}, which means that no direct eye information is available for eye reconstruction in this branch), and mostly relies on the gaze-related representation injected from the full-face branch, thus pushing the full-face branch to maximally promote its gaze feature encoding ability. In this work, we utilize HRNet~\cite{sun2019high} to detect the eye region. In particular, we use the AFLW-pretrained model which outputs $19$ facial landmarks for each input face image. Among the $19$ landmarks, we select a total of four landmarks representing respectively the left/right corners of both eyes. We consider the midpoint of the line connecting the left and right corners of the eyes to be the eye center, and select the $3\times 4$ patches around this eye center to serve as eye-related patches.

To propagate information across both branches, an injection bottleneck structure is employed, which first reduces the dimensionality of the output of the full-face feature decoder via an average pooling layer to pool all patches into one patch and a linear layer to reduce its dimension to $16$, and then performs feature up-sampling to restore the high dimensional feature to match the dimensionality of the decoder in the eye-masked branch by another linear layer. The up-sampled feature is injected to the eye-masked branch as an extra patch-level token representation by stacking. The basic idea of using this injection bottleneck structure is that by dimensionality reduction, global information about gaze direction could be discovered/preserved and those detailed local textural information from eye pixels are attenuated. As can be seen from Fig.~\ref{fig:motivation}, the proposed injection bottleneck
helps the eye masked auto-encoders to reconstruct the eye patches with correct gaze directions, indicating that we can avoid simple eye pixel replication and detailed eye texture learning.
We use the conventional mean square error (MSE) for the masked eye reconstruction loss function, \emph{i.e.}, between the pixels of the reconstructed eye related patch and the masked ground-truth, expressed as,
\begin{equation}
\mathcal{L}_{rec}=|\mathbf{I}_e^{pred}- \mathbf{I}_e^{gt}|,
\end{equation}
where $\mathbf{I}_e^{pred}$ and $\mathbf{I}_e^{gt}$ denote the reconstructed and ground-truth eye patch pixels, respectively.

\textbf{Gaze Information Squeeze.} Despite the above efforts in structural design, the full-face feature encoding branch still has the tendency and possibility to learn facial appearances/textures instead of gaze direction information.
Therefore, we propose a self-squeeze training scheme to further force the model to capture the gaze-related information from face patches while ensuring the reconstruction still pays attention to the gaze direction rather than the facial textures, which includes two designs, namely, 1) \emph{asymmetric encoder-decoder} and 2) \emph{weight sharing mechanism}.
First, different from the conventional asymmetric design in MAE where the observed part of the signals are NOT masked out, in our asymmetric encoder-decoder, we mask not only the eye patches but also some facial patches in the masked encoder while reconstructing just the eye masked patches to ensure the injection bottleneck pays more attention to the gaze direction rather than the facial textures. Moreover, we have selected a high mask ratio of $71.6\%$ for the facial region (the total mask ratio for eyes and facial region is $75\%$) to encourage the injection bottleneck to encode more gaze-related information that is not limited to the eye area, \emph{i.e.}, head pose to help the eye feature reconstruction and enable our EM-IB to infer gaze with the high-variance head pose.
Second, as mentioned above, the weight-sharing mechanism employs the same vision transformer for both eye-masked encoder as full-face injection encoder. The reason is as follows.
The gradients of the eye reconstruction loss are propagated to the network through both branches, which could be denoted as,
\begin{equation}
\mathbf{g}_{\theta} = \frac{\partial{\mathcal{L}}}{\partial{\mathbf{z}_b}} \frac{\partial{\mathbf{z}_b}}{\partial{\boldsymbol{\theta}}} + \frac{\partial{\mathcal{L}}}{\partial{\mathbf{z}_o}} \frac{\partial{\mathbf{z}_o}}{\partial{\boldsymbol{\theta}}} = \mathbf{g}_{inj} + \mathbf{g}_{rec},
\end{equation}
where $\boldsymbol{\theta}$ is the parameters of the encoder (same for both branches as in weight sharing scheme), and $\mathbf{z}_b$ and $\mathbf{z}_o$ stand for the injection vector and encoded token of the reconstruction branch, respectively. Namely, the gradients of the encoder ($\mathbf{g}_\theta$) consist of both gradient components from the injection branch ($\mathbf{g}_{inj}$) as well as the reconstruction branch ($\mathbf{g}_{rec}$). Therefore, in order to achieve better eye reconstruction, the network would also look into the feature representation provided by the full-face branch, which enforces the full-face branch to extract useful eye-related information. The combination of both designs distill the position embedding into the injection encoder and force the injection encoder to learn the head pose (gaze-related) information that cannot be accessed directly.

\textbf{Eye/gaze Information Contrastive Loss.} Although full-face image provides all eye related information, this information is surrounded/confused/coupled by non-eye features, in a dimension reduced space (\emph{i.e.}, through feature learning). Therefore, if too much reliance is placed on extracting eye information from full-face images to reconstruct the full details of the masked eye region, the model will tend to extract over-fitting feature representations from non-eye regions. To avoid this problem, we propose an eye/gaze information contrastive learning objective to compliment the masked reconstruction loss, in the unsupervised setting.
Our idea is as follows. The eye patch provides all fine details of eye self-reconstruction (\emph{i.e.}, could be regarded as positive information); the full face image also provides the overall information about the eyes, but due to feature learning and abstraction, it inevitably loses some detailed information (\emph{i.e.}, could be considered as neutral information); and the eye/gaze related information contained in the face image with eye region masked-out is extremely limited, and this information might even interfere with the true eye feature (\emph{i.e.},  thus could be considered as negative information).
According to the above facts, when the feature extractor learns the appropriate eye representation, it should possess the following property: the eye patch reconstruction error with the input of positive information must be smaller than that with the negative information. In other words, we constrain our feature learning module to not \textbf{over-extract} information from the full-face image to reconstruct the eye details, which easily leads to over-fitting.
More specific, we define the following feature triplets: 1) the single full-face image feature (denoted as $\mathbf{f}^{0}$), 2) the fused feature composed of full-face image and eye patches (denoted as $(\mathbf{f}^{+}, \mathbf{f}^{0})$), and 3) the fused feature composed of full-face image and eye masked-out regions (denoted as $(\mathbf{f}^{-}, \mathbf{f}^{0})$).
Then, the above constraint can be expressed mathematically as
\begin{equation}
\mathcal{L}_{contr}=\max\big(|\mathcal{F}(\mathbf{f}^{+},\mathbf{f}^{0})- \mathbf{I}_e^{gt}|-|\mathcal{F}(\mathbf{f}^{-},\mathbf{f}^{0})- \mathbf{I}_e^{gt}|,0\big),
\end{equation}
where $\mathcal{F}(.)$ denotes the feature fusion/combination network.
Note that this triplet based loss penalizes the situation when the reconstructed eye patch based on negative information is \emph{more accurate} than that based on positive information(\emph{i.e.}, which is not reasonable indicating over-fitted feature). Thus, in par with this auxiliary learning/regularization loss, reasonable eye/gaze related feature could be learned from the full-face image and the potential over-fitting caused by attending \emph{too much} on the non-eye region can be effectively alleviated, during the unsupervised pre-training phase.
The overall unsupervised pre-training loss could be thus summarized as,
\begin{equation}
\mathcal{L}=\mathcal{L}_{rec} + \lambda\mathcal{L}_{contr},
\end{equation}
where the weighting factor $\lambda$ is set empirically as $0.01$ in this work.

\subsection{Train CNNs with Knowledge Distillation} \label{sec:distill_meth}
To show the general applicability of our EM-IB scheme, we also train CNN based full-face unsupervised gaze representation.
To do so, a straightforward way is to take the ViT model pre-trained with EM-IB as the upper MAE branch, and use a ResNet based CNN model as injection encoder. However, such design might result in seriously degraded training performance, since the above proposed weight-sharing training strategy is no longer applicable, \emph{i.e.}, shared Vit structures for both full-face and eye masked reconstruction branch.
Therefore, here we propose to train CNNs with knowledge distillation, \emph{i.e.}, to distill the knowledge from EM-IB pre-trained ViT to ResNet.
More specific, to train CNN-based full-face feature representation (injection bottleneck feature), we employ a ViT based teacher model sharing the same weight with the upper MAE branch. A student model based on ResNet is then learned by imposing a mean square error (MSE) based distillation loss between teacher (ViT) and student (ResNet). Thus the total loss during EM-IB training includes this distillation component.
In experiments, we decrease the distillation loss gradually during training, and we observe that ResNet could get similar performance compared with the ViT teacher model.

\section{Experiments}
We perform the following experiments to comprehensively evaluate the effectiveness of the proposed unsupervised pre-training scheme EM-IB for gaze estimation. These include 1) a two-stage pre-training on ImageNet~\cite{imagenet_cvpr09} as well as large scale face/gaze datasets (including the commonly used face repository GazeCapture~\cite{gazecapture}) and linear probing on various gaze benchmarks with comparisons to the state-of-the-art (at both supervised and unsupervised settings); 2) a few shot calibration learning with both within- and cross-dataset settings to evaluate the validity/generalization capability of our pre-training scheme; 3) fine-tuning experiments to further demonstrate the advantage of the pre-trained representation; and 4) visualizations of the working mechanism of the proposed information injection design; as well as 5) a series of ablation studies on key parameters such as mask ratio, number of calibration samples, and the proposed network structural settings, \emph{etc}.

\subsection{Experiments Setup} \label{sec:exp_setup}

\paragraph{Datasets} We employ $5$ gaze datasets for evaluation: Columbia Gaze~\cite{CAVE_0324}, MPIIFaceGaze~\cite{zhang2017s}, EyeDiap~\cite{funes2014eyediap}, Gaze360~\cite{kellnhofer2019gaze360} and ETH-Xgaze~\cite{zhang2020eth}.

\textbf{ColumbiaGaze} (Columbia) consists of $5.8K$ images with five horizontal head poses from $56$ subjects.

\textbf{MPIIFaceGaze} (MPII) contains $45K$ images from $15$ subjects collected in front of laptops.

\textbf{EyeDiap} contains video clips from $16$ subjects and screen targets or $3D$ floating balls. We follow ~\cite{yu2020unsupervised} and select $5$ sessions with HD video, condition B, floating target and static head pose for experiments.

\textbf{Gaze360} contains $85K$ images from $238$ subjects with a wide-range of head poses and gaze directions. We only evaluate the method with the frontal faces like ~\cite{zhang2020eth}.

\textbf{ETH-Xgaze} (Xgaze) contains $756K$ images in laboratory environments with large variations in head poses and gaze directions.

We follow ~\cite{zhang2018revisiting}, ~\cite{cheng2021appearance} to perform data normalization on all facial images with detected landmarks~\cite{wang2020deep}.
All the experiments are conducted in a cross-person setting, and only unlabeled images in training sets are used for EM-IB pre-training.
Note that we also use the large scale face-gaze dataset \textbf{GazeCapture} for a second phase pre-training (\emph{i.e.}, after first phase ImageNet general image pre-training) and compare the calibrated performance on the evaluation datasets, to demonstrate the maximum ability of our scheme, by comparing with both state-of-the-art unsupervised and fully supervised methods. Note that GazeCapture is a very large face-gaze repository suitable for pre-training which contains facial/eye data from over $1450$ people consisting of almost $2.5M$ frames.
\paragraph{Evaluation Settings}
5-folds evaluation  protocol is used for ColumbiaGaze and leave-one-out protocol is used for MPIIGaze and EyeDiap. The official train-test splitting manner is adopted for Xgaze and Gaze360.
For each fold, we first pre-train the network with EM-IB within the split train set and adapt it to randomly selected labeled samples in the train set.
The \textbf{average angular gaze error} (in degree$^\circ$) is used as the evaluation metric, which can be computed by calculating the angle between two gaze directions.
\paragraph{Implementation Details} 
We take both original vision transformer (ViT) (ViT-base) as in~\cite{dosovitskiy2020image} and a smaller ViT (ViT-tiny) as in~\cite{pmlr-v139-touvron21a} as encoder/decoders, accordingly to different experimental settings.
All of the input images are resized to $224\times 224$, and split into $16\times16$ patch tokens.
During the pre-training period, we take ViT-tiny as backbone and share the weights between two branches of EM-IB if without extra notes. We use detected facial landmarks~\cite{wang2020deep} to decide the masked eye patches and the remaining facial patches are masked randomly. For the reconstruction branch, we set the total masking ratio to $75\%$ and both eyes are masked. For the information injection branch, we mask one or two eyes with $50\%$ probability to enhance the learning robustness of the network. For the injection bottleneck, we set the injection dimension as $16$ to balance the amount of injection information and the novel information from the few-shot gaze adaptation.
All the models are initialized from MAE~\cite{he2022masked} pre-trained weights on ImageNet~\cite{imagenet_cvpr09} first, and then pre-trained further on large scale face datasets (\emph{i.e.}, GazeCapture).
For ColumbiaGaze and Eyediap, the scale of data is too small ($5.8K$ images and $12K$ frames of video for ColumbiaGaze and Eyediap respectively), so we pre-train the model on Xgaze.
Also note the except for the main experiment (\emph{i.e.}, linear probing), the contrastive loss term is unset for training efficiency.

\subsection{Results of Linear Evaluation (w/o calibration) with the Whole Dataset} \label{sec:whole_linear-result}
\begin{table}
\renewcommand{\tabcolsep}{1pt}
\centering
\caption{Results of linear probing with the whole datasets for cross-person gaze estimation (without calibration). The backbone is ViT-tiny.}\label{tab:linear_full}

\begin{tabular}{p{2.3cm}<{}p{2.1cm}<{\centering}p{2.1cm}<{\centering}p{2.1cm}<{\centering}}
\toprule
Methods &  MPII  & Xgaze & Gaze360 \\
\midrule
SimCLR~\cite{simclr}  & 9.28 & 26.57  & 30.79 \\
MAE~\cite{he2022masked} &7.12 &17.68 & 19.47 \\
MAE-single  &6.78    & 13.06    & 15.65  \\
AE   & 6.36 & 12.72 & 14.40 \\
EM-IB (192 dim)   &  5.19 & 11.25  & 14.05    \\
EM-IB  (16 dim) &  \textbf{4.58}     & \textbf{10.47}   & \textbf{13.13}    \\

\midrule
MAE-single (CapGaze-pretrain)   &   5.5   & 14.52 &  14.03\\
AE (GapGaze-pretrain)    &    4.54   & 11.24 &  13.42 \\
EM-IB  (CapGaze-pretrain)  &   \textbf{4.18}  &  \textbf{9.72} & \textbf{12.74}  \\

\midrule
MAE (MLP-2/3/4)   &   6.18/6.02/6.19   &  16.18/16.02/15.94  &  14.34/14.61/14.56 \\
AE (MLP-2/3/4)    &    4.74/4.71/4.81   &   11.39/11.37/11.34 &   13.79/13.72/13.72\\
EM-IB (MLP-2/3/4)  &  \textbf{4.65/4.50/4.55}   &  \textbf{10.55/10.50/10.65} & \textbf{13.13/13.12/13.15}  \\
\midrule
FullFace~\cite{zhang2017s} (Supervised)   &    4.93  &  7.38  &  14.99 \\
RT-Gene~\cite{fischer2018rt} (Supervised)   &    4.66  &  N/A  &  12.26 \\
Dilated-Net~\cite{chen2018appearance} (Supervised)   &   4.42   &  N/A  &  13.73 \\
GazeTR-Hybrid~\cite{cheng2022gazetr} (Supervised)   &    4.00  &  N/A  &  10.62 \\
\bottomrule

\end{tabular}  
\end{table}


 \label{sec:linear_evaluation}
We first evaluate the unsupervised gaze representation learned by EM-IB with linear probing with the whole dataset (\emph{i.e., without dataset calibration}) following the convention of self-supervised learning methods. As the low dimensional injection bottleneck is designed for few-shot learning and other baseline methods like SimCLR~\cite{simclr} contain no bottleneck, the whole-dataset linear evaluation is fairer. The backbone is ViT-tiny with a feature dimension as $192$.
To maintain fairness, all the methods are required to output a $192$-dimensional feature after an unsupervised learning phase for evaluation. Our proposed EM-IB contains a $192-16-192$ bottleneck, so we use the first $192$-dim feature before down-sampling for evaluation, and mark this method as EM-IB ($16$ dim). To evaluate the effectiveness of the information bottleneck design, we add a variant of EM-IB that contains no bottleneck, denoted as EM-IB ($192$ dim), which only utilizes a $192-to-192$ linear transform.
We compare EM-IB with contrastive learning method SimCLR~\cite{simclr}, MAE~\cite{he2022masked}, MAE-single, Auto-Encoder (AE) and EM-IB without injection bottleneck. The settings of SimCLR and MAE are the same as their original papers, and MAE-single denotes MAE with a specific mask scheme (only mask one eye and other facial patches). For each dataset, we train a linear regressor upon the learned gaze feature with the whole dataset (on train split). The linear evaluation results in Tab.~\ref{tab:linear_full} show that the representation learned by EM-IB contains more gaze information compared to other methods. Specifically, EM-IB with $16$-dimensional injection bottleneck outperforms EM-IB ($192$ dim), which proves that the designed bottleneck not only benefits few-shot learning but also helps to squeeze gaze information to the decoder and prevents pixels being directly copied from the injection branch.

To test whether the pre-trained representation has good adaptation ability, we also replace the linear probing component (\emph{i.e.}, linear regressor) with a multi-layer MLP component (including two, three and four layers denoted as $MLP-2$, $MLP-3$ and $MLP-4$, respectively), for the comparing algorithms. As shown in Tab.~\ref{tab:linear_full}, after adaptation training by MLP layers, most comparing algorithms present better accuracy than using linear probing (\emph{i.e.}, $MLP-1$), which implies that all these algorithms have representation adaption capability. Also, our proposed EM-IB representation still outperforms other methods.

In addition, we also conduct pre-training experiment on the large scale face-gaze dataset GazeCapture~\cite{gazecapture}, to further test the potential of unsupervised pre-training schemes (specifically on face-gaze big data) for both our method and the comparing ones (note that pre-training on ImageNet is still performed before pre-training on GazeCapture, for feature generalization). For benchmarking, we also report the performances on the state-of-the-art fully-supervised method FullFace~\cite{zhang2017s}, RT-Gene~\cite{fischer2018rt}, Dilated-Net~\cite{chen2018appearance}, GazeTR-Hybrid~\cite{cheng2022gazetr}. For unsupervised methods, linear probing is adopted. As shown in Tab.~\ref{tab:linear_full}, we see that based on GazeCapture pre-training, all comparing methods including ours achieve better linear probing results, compared to the pre-training results on ImageNet only, which implies that information extracted from large scale face-gaze dataset is beneficial for more generalized and more accurate gaze representation. Also, our EM-IB still performs the best which demonstrates its good representation capability. Finally, we note that the performance of EM-IB is comparable with (or even better than) most state-of-the-art fully-supervised method (\emph{i.e.}, only inferior to most recent strong estimator GazeTR-Hybrid~\cite{cheng2022gazetr} which specifically fine-tunes the transformer parameters using supervised training data), which indicates its great application potential.


\subsection{Results of Few-shot Calibration Gaze Estimation} \label{sec:few-shot-result}
Following~\cite{yu2020unsupervised} and~\cite{sun2021cross}, we evaluate the unsupervised performance of EM-IB with $100$-shot linear probing experiments, under both within-dataset and cross-dataset settings. We report average angular gaze error with standard deviation over 3 repeated experiments.
\begin{table*}[!htbp]
\renewcommand{\baselinestretch}{1.0}
\renewcommand{\arraystretch}{1.0}
\setlength{\tabcolsep}{1pt}
\caption{Results of 100-shot gaze estimation for cross-person estimation. We compare with Yu \protect~\cite{yu2020unsupervised} and Cross Encoder (CE) \protect~\cite{sun2021cross}. The ResNet18 version EM-IB is trained with pre-trained ViT-tiny as teacher model. Angular gaze error ($^\circ$) (mean$\pm$std) is reported. Each data pair reports the result without (left) and with (right) head pose.}
\centering
\begin{tabular}{lc<{\centering}p{1.3cm}<{\centering}p{1.35cm}
    <{\centering}p{1.3cm}<{\centering}p{1.3cm}<{\centering}p{1.3cm}<{\centering}p{1.3cm}<{\centering}p{1.3cm}<{\centering}p{1.3cm}<{\centering}p{1.3cm}<{\centering}p{1.3cm}}
\toprule
\multirow{2}*{\textbf{Methods}}  & \multirow{2}*{Backbone}  & \multicolumn{2}{c}{\textbf{Columbia}} & \multicolumn{2}{c}{\textbf{MPII}} &\multicolumn{2}{c}{ \textbf{EyeDiap}} & \multicolumn{2}{c}{\textbf{Xgaze}} & \multicolumn{2}{c}{ \textbf{Gaze360}} \\
  &   &  w/o & w/ head  & w/o & w/ head&  w/o & w/ head & w/o & w/ head &  w/o & w/ head \\


\midrule
Yu et.al~\cite{yu2020unsupervised} & ResNet-style  & -    & 8.95  & -   & - & - & 9.02   & - & -     & -      & -    \\
CE~\cite{sun2021cross}   & ResNet18 & 7.1\tiny{$\pm0.1$} & 6.4\tiny{$\pm0.1$}
& 7.2\tiny{$\pm0.2$} & 7.2\tiny{$\pm0.2$}  & -     & -      & -   & -     & -      & -    \\
\midrule
Auto-Encoder & ViT-tiny  & 9.4\tiny{$\pm0.1$} & 9.4\tiny$\pm0.1$   &
6.8\tiny{$\pm0.2$} & 6.5\tiny{$\pm0.2$}
& 10.3\tiny{$\pm0.4$} & 12.9\tiny{$\pm0.6$}
&  19.3\tiny{$\pm0.5$} & 16.8\tiny{$\pm0.6$}
& 24.3\tiny{$\pm0.4$}  & 18.6\tiny{$\pm0.3$} \\
MAE & ViT-tiny &   12.5\tiny{$\pm0.1$} & 12.3{\tiny$\pm0.2$}   &
8.8\tiny{$\pm0.1$} &8.9\tiny{$\pm0.1$}
& 26.2\tiny{$\pm0.5$} & 23.8\tiny{$\pm0.7$}
&  31.1\tiny{$\pm0.6$} & 28.3\tiny{$\pm0.6$}
& 37.1\tiny{$\pm0.5$}  & 24.0\tiny{$\pm0.5$} \\

MAE-single & ViT-tiny &   12.1\tiny{$\pm0.2$} & 12.2{\tiny$\pm0.2$}   &
8.7\tiny{$\pm0.2$}&  8.8\tiny{$\pm0.2$}
& 25.5\tiny{$\pm0.4$} & 24.4\tiny{$\pm0.4$}
&  26.0\tiny{$\pm0.4$}& 23.6\tiny{$\pm0.3$}
& 22.2\tiny{$\pm0.4$}  & 17.0\tiny{$\pm0.5$} \\
\midrule
EM-IB (Ours) & ViT-tiny  &  6.1\tiny{$\pm0.1$}  & 6.1\tiny{$\pm0.1$}
& 5.5\tiny{$\pm0.1$}& 5.1\tiny{$\pm0.1$}
& 7.6\tiny{$\pm0.2$}& 7.8\tiny{$\pm0.2$}
& 14.4\tiny{$\pm0.3$} &11.8\tiny{$\pm0.1$}
&15.6\tiny{$\pm0.3$} &15.1\tiny{$\pm0.3$} \\

EM-IB (Ours) & ResNet18  &  6.1\tiny{$\pm0.1$}  & 6.1\tiny{$\pm0.1$}
& 5.6\tiny{$\pm0.1$}& 5.2\tiny{$\pm0.1$}
& 7.6\tiny{$\pm0.2$}& 7.6\tiny{$\pm0.2$}
& 14.1\tiny{$\pm0.3$} & 11.7\tiny{$\pm0.2$}
&15.5\tiny{$\pm0.3$} &15.0\tiny{$\pm0.3$} \\

EM-IB (Ours) & ViT-base  & \textbf{5.8}\tiny{$\pm0.1$} & 5.8\tiny{$\pm0.1$}
& 5.1\tiny{$\pm0.1$} & \textbf{5.1}\tiny{$\pm0.1$}
& \textbf{7.3}\tiny{$\pm0.1$} & 7.4\tiny{$\pm0.1$}
& 13.5\tiny{$\pm0.2$} & \textbf{11.5}\tiny{$\pm0.3$}
& 14.7\tiny{$\pm0.2$} & \textbf{14.6}\tiny{$\pm0.2$}  \\

\bottomrule

\end{tabular}

\label{tab:100shot}
\end{table*}

\subsubsection{Within-dataset Few-shot Evaluation}
After unsupervised pre-training with EM-IB on Columbia, MPII, EyeDiap, Gaze360, and Xgaze respectively, we conduct few-shot gaze adaptation by training a simple linear layer for within-dataset evaluation, and the results are exhibited in Tab.~\ref{tab:100shot}. Note that this linear layer has only $16 \times 2 + 2 = 34$ parameters (without head pose), which makes it possible to train on only $100$ samples. We exhibit the results of EM-IB with a backbone of ViT-tiny, ViT-base and ResNet-18 in Tab.~\ref{tab:100shot}. For ResNet-18, we take pre-trained ViT-tiny as teacher model and help it train with a distillation loss as elaborated in the methodology section. The results in Tab.~\ref{tab:100shot} show that with the help of pre-trained ViT-tiny, ResNet-18 could get equivalent or even better results. This implies our EM-IB method could be expanded to CNNs with proposed distillation strategy. Similarly, results on ViT-base indicate that EM-IB could be expanded to a larger model for better performance.

\paragraph{Comparing with $3$ Baseline Methods} Auto-Encoder, MAE and MAE-single. The \textbf{Auto-Encoder} method (intermediate feature dimension = $16$) is equivalent to setting the masking ratio of the first branch to $100\%$ in EM-IB. In this case, the reconstruction branch only provides the position embedding while the injection branch provides the feature learned from the full-face input. The \textbf{MAE}~\cite{he2022masked} method reconstructs images from $75\%$ random masked inputs. The \textbf{MAE-single} method, with eye-specific masking scheme, masks one eye of input face (and a total of $75\%$  of the other facial patches) and attempts to reconstruct the eyes. To accommodate with the few-shot linear probing setting, we add an extra $16$-dimensional feature bottleneck in MAE and MAE-single to concatenate with other encoded features for decoding as we do for EM-IB.
The results in Tab.~\ref{tab:100shot} show that the auto-encoder performs best among 3 baselines in $100$-shot linear-probing evaluation. This result is reasonable because both MAE and MAE-single do not have complete face as input and lack a reasonable design of feature bottleneck.
The ViT-tiny version of EM-IB outperforms $3$ baselines with large margin in the within-dataset $100$-shot linear probing, \emph{i.e.}, outperforms the auto-encoder baseline on $5$ benchmarks by $35.1\%$, $19.1\%$, $26.2\%$, $25.4\%$ and $35.8\%$, respectively.

\paragraph{Comparing with the State-of-the-Arts}  We then compare with two unsupervised gaze estimation methods~\cite{yu2020unsupervised, sun2021cross}. They are both trained on eye patches with ResNet-style backbone. We emphasize that our proposed EM-IB is designed for ViT-style network, but we could distill the model to CNN as we show in the extensive distillation experiments. The design of EM-IB introduces full-face based gaze information input to unsupervised gaze learning, and improves the performance.
We see that even ViT-tiny outperforms all the state-of-the-art methods, \emph{i.e.}, gaze errors are reduced by $4.7\%$, $29.2\%$, and $15.7\%$ in Columbia, MPIIGaze, and EyeDiap, respectively. As ViT-tiny has only $4.2 M$ parameters (vs. ResNet $11.9M$), it is relatively fair to compare with ResNet-18 methods. Besides, the ResNet18 version of EM-IB with ViT-tiny as teacher model also outperforms these two methods.
Note that there is no unsupervised benchmark for datasets ETH-Xgaze and Gaze360 with a larger variance of gaze and head pose, and our EM-IB gives the first unsupervised benchmark for these large-variation gaze datasets.
The best result of ~\cite{sun2021cross} is achieved by concatenating with head pose and ~\cite{yu2020unsupervised} considers the gaze in head pose coordinating system directly. We also report the results with and without head pose concatenated, and results show that EM-IB performs well in few-shot gaze estimation without extra head pose. It proves that the gaze representation learned by EM-IB encodes the rotational information of both eyeball and head effectively.
\subsubsection{Cross-dataset Few-shot Evaluation}
To evaluate the domain adaptation ability of EM-IB, we conduct few-shot gaze estimation experiments in a cross-dataset manner and exhibit the results in Tab.~\ref{tab:cross100shot}. We take ETH-Xgaze as the source domain dataset, while Columbia, MPIIGaze, and Gaze360 as target domain datasets. The backbone is ViT-tiny. We first pre-train EM-IB on ETH-XGaze without labels, and then adapt the network with $100$ labeled data on the target domain dataset by linear probing. As head pose is not crucial for EM-IB, we report the results without using it.
\begin{table}
\renewcommand{\baselinestretch}{1.0}
\renewcommand{\arraystretch}{1.0}
\setlength{\tabcolsep}{1pt}
\centering
\caption{Results of cross-dataset $100$-shot gaze estimation. The source domain dataset is Xgaze.  }\label{tab:cross100shot}
\centering
\begin{tabular}{p{2.9cm}<{}p{1.4cm}<{\centering}p{1.3cm}<{\centering}p{1.1cm}<{\centering}p{1.2cm}<{\centering}}
\toprule
Methods    & Backbone & Columbia & MPII  & Gaze360 \\
\midrule
Supervised~\cite{zhang2020eth} & ViT-tiny  & \textbf{5.89} & \textbf{5.21} & 18.03          \\
\midrule
CE~\cite{sun2021cross} & ResNet18    &7.76 & 9.04 & -      \\
EM-IB & ViT-tiny   & 7.23  & 6.44   & 17.21  \\
\bottomrule
\end{tabular}
\end{table}
We compare EM-IB with both the supervised method~\cite{zhang2020eth} and the unsupervised method CE~\cite{sun2021cross}. For the supervised method, we add a $16-to-2$ FC layer to ViT-tiny as the last layer, train it on Xgaze with supervision, and get a gaze error of $4.37$ on the Xgaze test set. Then we train the last $16$-dim FC layer on the target dataset with $100$ samples as we did for EM-IB. \textbf{We find that EM-IB even outperforms the supervised method on Gaze360 when adapting from Xgaze.} It indicates that unsupervised features learned by EM-IB perform better in cross-domain datasets with a wide-range gaze distribution. As ViT-tiny contains less parameters than ResNet-18, we use the ViT-tiny version of EM-IB to compare with CE~\cite{sun2021cross}, and our proposed EM-IB also consistently outperforms CE on both Columbia and MPIIGaze.

\subsection{Results of Fine-tuning with Subset of Dataset}
\begin{figure}[!t]
\renewcommand{\baselinestretch}{1.0}
\centering
\includegraphics[width=1.0\linewidth]{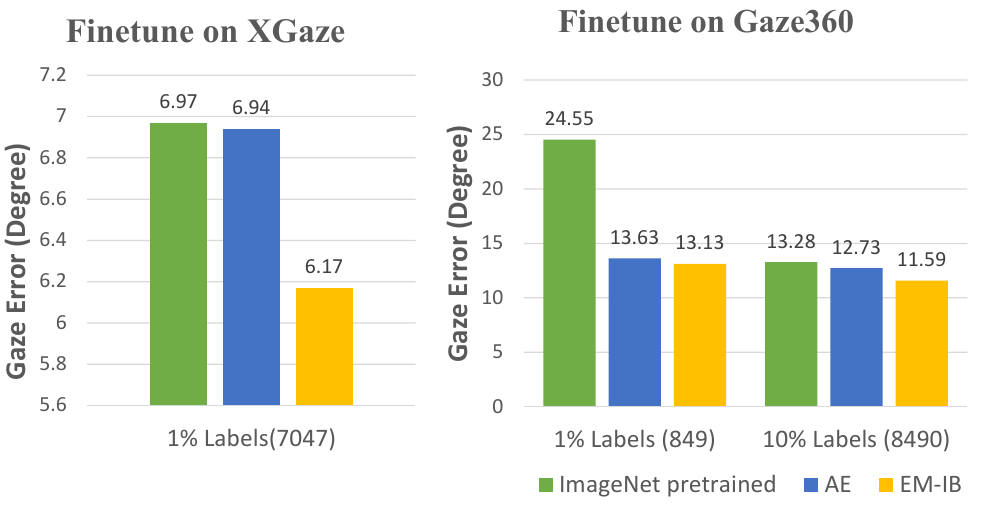}
 \caption{Results of fine-tuning the model with the subset of Gaze360 and XGaze. The backbone is ViT-tiny.
}   \label{fig:finetune_subset}

\end{figure}

We further evaluate the generalization capability of the learned gaze representation by fine-tuning the model with a subset of the dataset. In this setting, we use random sampled subset of dataset to fine-tune the unsupervised pre-trained ViT model. We compare our EM-IB pre-trained model with MAE pre-trained model on ImageNet and auto-encoder (AE) pre-trained model. We evaluate on two challenging gaze datasets: Xgaze and Gaze360, and all comparing methods are further fine-tuned on Xgaze and Gaze360, respectively. The results are illustrated in Fig.~\ref{fig:finetune_subset} and it shows that our proposed EM-IB improves the gaze estimation accuracy when data is not sufficient. In particular, when fine-tuning with only $1\%$ data of Gaze360, the MAE pre-trained model on ImageNet could not generalize well, while EM-IB pre-trained model gets a $46.5\%$ improvement over the baseline.

\begin{figure*}[htb]
\renewcommand{\baselinestretch}{1.0}
\centering
\includegraphics[width=1.0\linewidth]{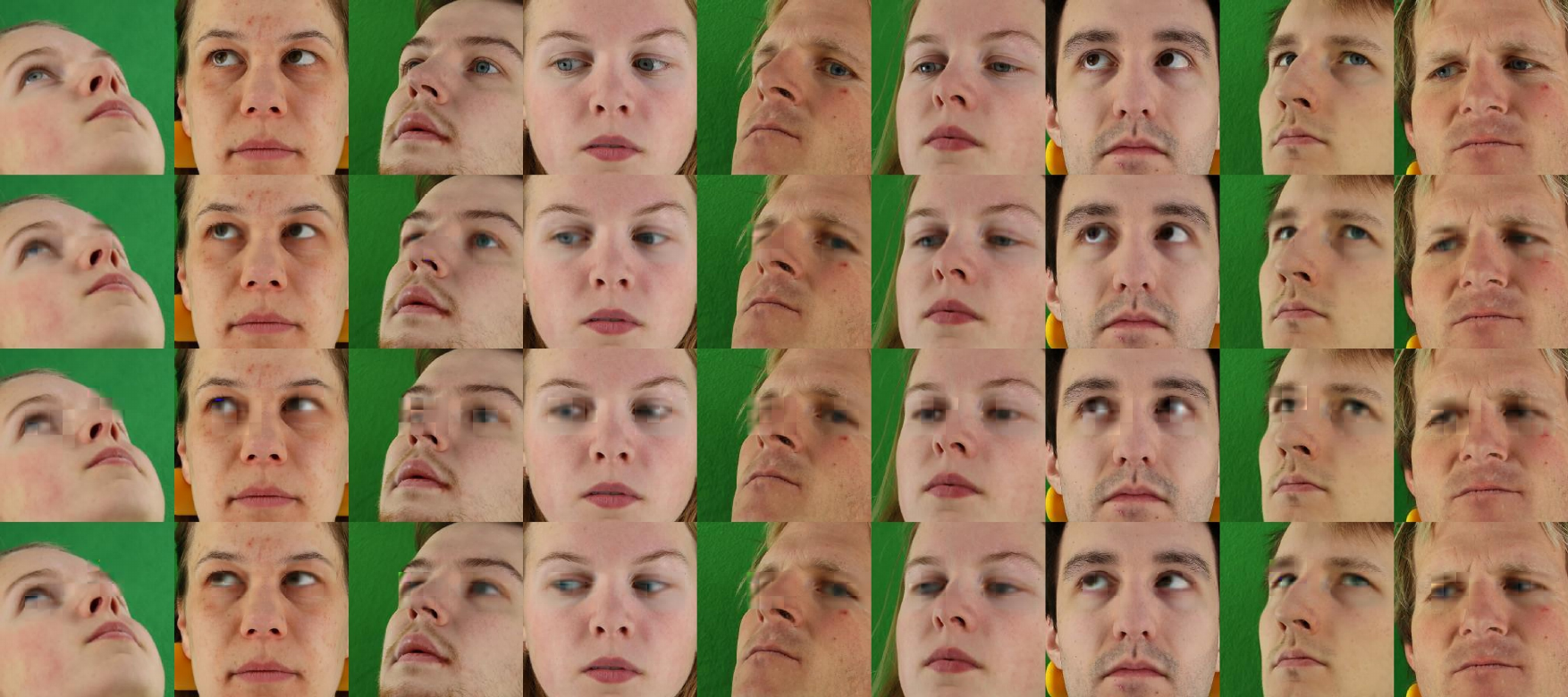}
 \caption{Visualization of gaze reconstructed eyes. For top to bottom rows, we show the 1) original ground-truth image, and results obtained by 2) our EM-IB, 3) AE and 4) MAE-single, respectively.
}    \label{fig:gaze-reconstruction}
\end{figure*}

\begin{figure}[htb]
\renewcommand{\baselinestretch}{1.0}
\centering
\includegraphics[width=1.0\linewidth]{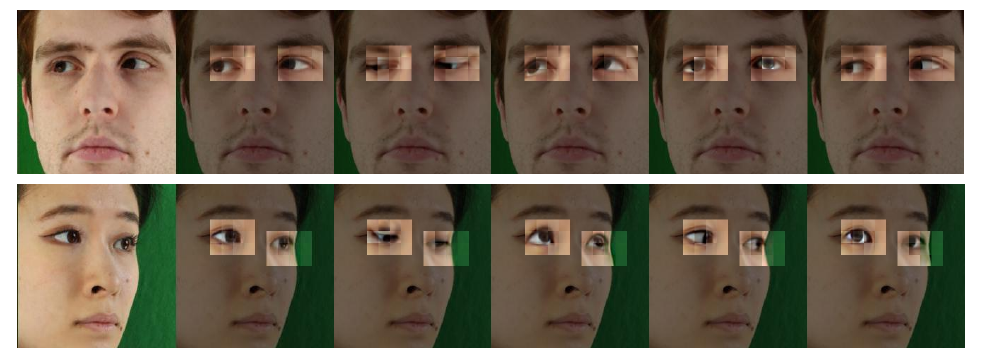}
 \caption{Visualization of gaze redirected eyes by our EM-IB method. For each row of images, we show the (1) original image, (2) reconstructed image, and (3-6) gaze redirected images where the person looks down, up, right, and left respectively compared to the original gaze.
}    \label{fig:gaze-redirection}
\end{figure}

\subsection{Visualization of Gaze Recontruction/Redirected Eyes}
To verify that the representation learned by EM-IB does contain abundant gaze information, we extent EM-IB for gaze reconstruction and redirection tasks.
Specifically, we can 1) reconstruct the masked eye patch or 2) redirect the gaze of the face image by manipulating the injection vector.
Eye/gaze reconstruction results are visualized in Fig.~\ref{fig:gaze-reconstruction}. While the reconstructed results obtained by our EM-IB present high accuracy, we note from Fig.~\ref{fig:gaze-reconstruction} that the results obtained by AE and MAE present averaged/blurred or incorrect eye/gaze reconstructions, which further demonstrate that 1) AE learns attenuated gaze representation from full-face while 2) MAE easily learns over-fitted gaze representation due to the lack of eye patch.
For redirection, first we utilize the weights of the learned linear layer to perturb the injection vector. We then inject the perturbed vector into the decoder and guide it to generate the eyes patches looking at up, down, left, and right directions, respectively.
We visualize the resulting images in Fig.~\ref{fig:gaze-redirection}. While the visualization is noisy due to the lack of an adversarial and perceptual loss, it is sufficient to tell that the gaze has been successfully redirected. It is proven that gaze information does propagate to the decoder through an injection bottleneck.

\subsection{Study on Eye/Gaze Information Contrastive Loss}
\begin{figure}[htb]
\centering
\includegraphics[width=1.0\linewidth]{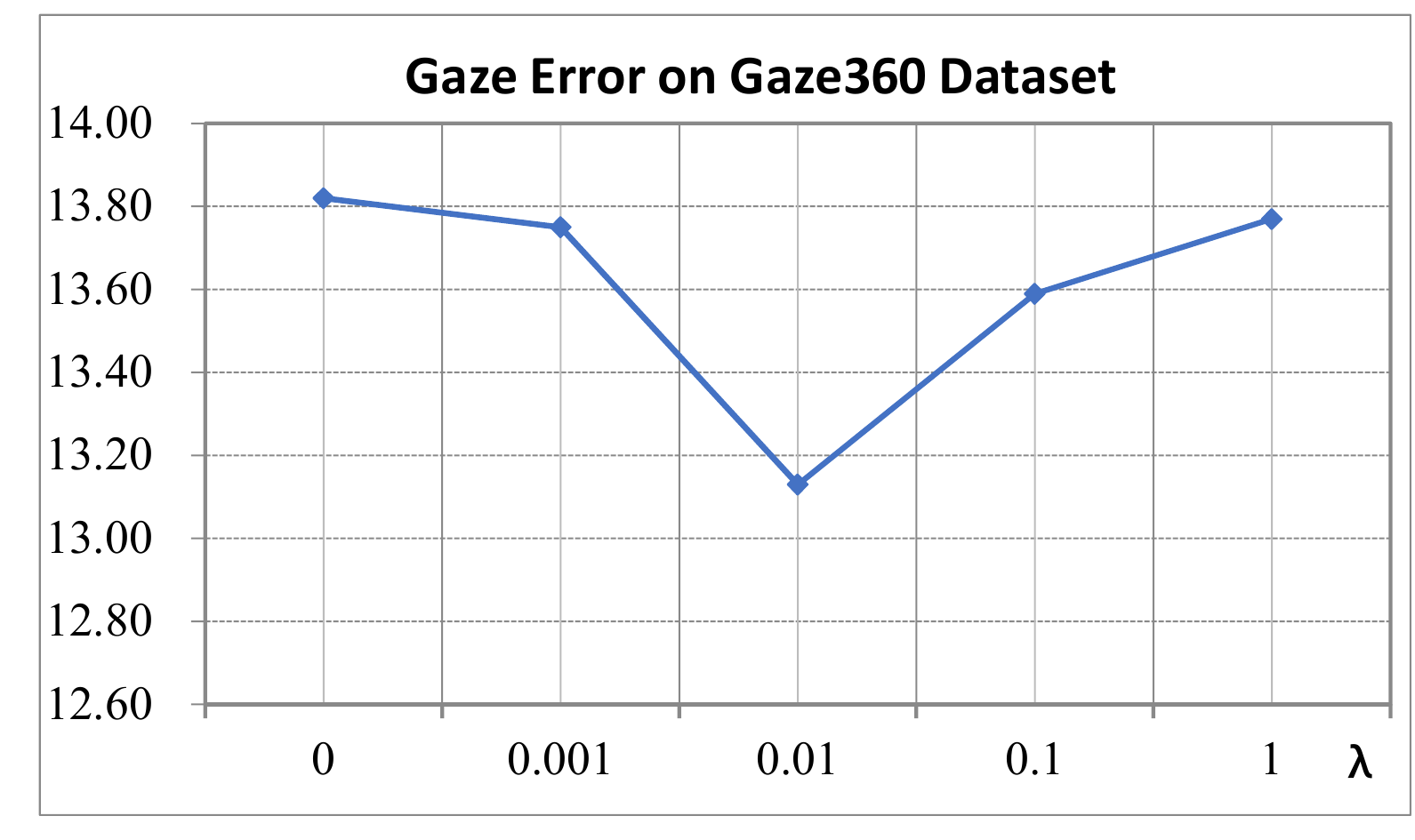}
\caption{Change of gaze estimation accuracy at different weights of the eye/gaze information contrastive loss.}
\label{fig:regulosseffect}
\end{figure}

Besides, in order to demonstrate the effectiveness of our proposed regularization constraint based on information contrasting, we perform the following experiment. We set the regularization loss coefficient to $\lambda = [0.0\ 0.001, 0.01, 0.1, 1.0]$ (\emph{i.e.,} here $0$ denotes without this regularizer) respectively, and observe the changes of the learned unsupervised gaze representation performance. We conduct this experiment on Gaze360, while results on other datasets present similar trend. Fig.~\ref{fig:regulosseffect} shows the results based on the whole dataset linear probing setup on Gaze360. We see that when the regularization coefficient increases from the minimum value, the representation performance gradually increases, and when it reaches a certain value of $0.01$, the gaze estimation performance reaches saturation and decreases slightly afterwards. This reflects that under a proper level of information contrasting, the model can better extract GAZE-related features from the full-face branch and avoid falling into an over-fitting state in the same time, and when the constraint becomes too strong, the reconstruction loss item is suppressed thus causing performance drop.

\subsection{Ablation Study}
\label{sec:ablation}




\paragraph{Ablation Study on Injection Dimension}
\begin{table}
\centering
\caption{Ablation study on injection dimension.} \label{tab:inj-dim}
\begin{tabular}{p{2cm}<{}p{1.2cm}<{\centering}p{1.3cm}<{\centering}p{1.2cm}<{\centering}}
\toprule
Injection Dim  & MPII & Gaze360 & Xgaze \\
\midrule
Dim = 8    &5.96  & 29.36   & 21.60 \\
Dim = 16  & \textbf{5.47}  & \textbf{14.92} & \textbf{14.36}  \\
Dim = 32  & 5.63 & 17.12  & 14.88  \\
\bottomrule
\end{tabular}
\end{table}
We evaluate the impact of various injection dimensions with a $100$-shot gaze estimation setting on MPIIGaze, Gaze360 and Xgaze.
The results in Tab.~\ref{tab:inj-dim} show that $16$ is the best choice (used in subsequent experiments) compared to $8$ and $32$ dimensions. The explanation could be as follows.
An injection bottleneck with a too-small feature dimension could hardly encode enough information like gaze, head pose, and other eye features for eye reconstruction. While an injection bottleneck with a large feature dimension makes it easy to over-fit the learned gaze representation, \emph{i.e.}, containing too much detailed face/eye textural or structural information.

\paragraph{Ablation Study on Sample Numbers of Calibration} In previous experiments we follow the convention~\cite{sun2021cross} to use $100$-shot adaptation to evaluate the performance of learned unsupervised feature. Here we evaluate the influence of sample numbers of few-shot gaze adaptation in Tab.~\ref{tab:sample numbers}. The more samples used for calibration, the lower the gaze error.
This indicates that our approach can be further improved with increasing number of labelled samples.
\begin{table}
\centering
\caption{Ablation study on sample numbers of calibration.}\label{tab:sample numbers}
\begin{tabular}{p{2cm}<{}p{1.2cm}<{\centering}p{1.3cm}<{\centering}p{1.2cm}<{\centering}}
\toprule
Sample Num & MPII & Gaze360 & Xgaze \\
\midrule
Num = 50  & 6.04 & 16.22 & 15.43 \\
Num = 100     &  5.47  & 14.92 & 14.36\\
Num = 200     & \textbf{5.31} & \textbf{14.49} &  \textbf{12.76} \\
\bottomrule
\end{tabular}
\end{table}

\begin{table}[!t]

\renewcommand{\tabcolsep}{1pt}
\caption{Ablation study on masking ratio (MR).} \label{tab:mask-ratio}
\centering
\begin{tabular}{p{1.8 cm}<{\centering}p{1.8cm}
<{\centering}p{1.4cm}<{\centering}p{1.6cm}<{\centering}p{1.6cm}<{\centering}}
\toprule
Masking ratio & with Head & MPII & Gaze360 & Xgaze \\
\midrule
12.5\% &  $\times$    &5.92  & 24.23   & 20.65   \\
12.5\%  & \checkmark  &5.33  & 17.69  & 14.22  \\
\midrule
50\% &  $\times$ & 5.45  & 16.21   & 16.38  \\
50\% &  \checkmark & \textbf{5.08}  & 15.96   & 12.34  \\
\midrule
75\%  & $\times${}   & 5.47 & 15.63 &  14.43 \\
75\%  & \checkmark{}    & 5.12 &  \textbf{15.12} &  \textbf{11.84} \\
\midrule
100\%  & $\times${}   & 6.77 & 24.25 & 19.27 \\
100\%  & \checkmark{}  &6.51  &18.62   & 16.80\\
\bottomrule
\end{tabular}
\end{table}
\paragraph{Ablation Study on Masking Ratio (MR)}
We change the masking ratio to verify the influence of the eye-related information squeeze strategy.
Illustrated in Tab.~\ref{tab:mask-ratio}, we evaluate the masking ratio (MR) of $12.5\%$, $50\%$, $75\%$ and $100\%$ on MPIIGaze, Gaze360, and ETH-Xgaze.
Note that the ratio of masking two eyes of the face image is $12.5\%$ and a vanilla auto-encoder indeed corresponds to a $100\%$ mask ratio.
Results show that a higher masking ratio on facial patches helps the network to encode more head pose information
in the injection bottleneck. When the masking ratio is equal to $12.5\%$, no extra facial patches are masked, and the
network relies heavily on extra head pose for gaze estimation, \emph{i.e.}, head pose reduces gaze error from $20.65$ to $14.22$ in Xgaze. When increasing the masking ratio to $75\%$, most facial patches are masked, encouraging the injection branch to encode more head pose information and lead to better gaze estimation. When we set masking ratio to $100\%$, the network degrades to an AE and both gaze-related and eye appearance correlated feature are encoded in a mixed way, thus harming the gaze estimation. This ablation study indicates that moderate ratio for the eye-masked encoder helps.

\begin{table}[!t]
\centering
\makeatletter\def\@captype{table}
\renewcommand{\baselinestretch}{1.0}
\renewcommand{\arraystretch}{1.0}
\setlength{\tabcolsep}{1pt}
\caption{Ablation study on weight sharing mechanism (WS). }
\label{tab:weight-sharing}
{
\begin{tabular}{p{2.0cm}<{\centering}p{1.0cm}<{\centering}p{1.0cm}<{\centering}p{1.0cm}<{\centering}p{1.0cm}<{\centering}p{1.0cm}<{\centering}p{1.0cm}}
\toprule
Weight-sharing  & \multicolumn{2}{c}{MPIIGaze} & \multicolumn{2}{c}{Gaze360} & \multicolumn{2}{c}{Xgaze} \\
&   w/o & w/head & w/o & w/head & w/o & w/head \\
\midrule
$\times$  & 7.45 & 6.82  & 35.93 & 24.12  & 16.62 & 11.94 \\
\checkmark   & 5.47 & \textbf{5.12} &15.63 & \textbf{15.12} & 14.43 & \textbf{11.84} \\
\bottomrule
\end{tabular}
}
\end{table}

\paragraph{Ablation Study on Weight Sharing Mechanism} We show the influence of weight-sharing mechanism in Tab.~\ref{tab:weight-sharing}.
We train two encoders separately (without the weight sharing mechanism), and then the gradients of the reconstruction branch have no impact on the injection branch. The $100$-shot linear-probing results are exhibited in Table.~\ref{tab:weight-sharing}. Results show this separated training strategy would seriously impair gaze representation learning and it indicates that the gradients of the reconstruction task also help the network training procedure. Besides, the weight-sharing mechanism is a computational-efficient and high-performance option.

\section{Conclusion}
In this paper, we propose a novel unsupervised gaze estimation framework, namely Eye Mask Driven Injection Bottleneck (EM-IB), which forces the full-face feature extraction module to learn a low dimensional gaze embedding without gaze annotations, through feature contrast and squeeze. Experimental results demonstrate that EM-IB significantly outperforms the state-of-the-art in unsupervised gaze estimation task.

\bibliographystyle{IEEEtran}

\newpage

\end{document}